\documentclass{article}
\usepackage{ijcai19}

\usepackage{times}
\usepackage{soul}
\usepackage{url}
\usepackage[utf8]{inputenc}
\usepackage[small]{caption}
\usepackage{graphicx}
\usepackage{amsmath}
\usepackage{booktabs}
\usepackage{algorithm}
\usepackage[noend]{algpseudocode}
\usepackage{helvet}  
\usepackage{courier}  
\usepackage{url}  
\usepackage{epsfig}
\usepackage{amssymb}
\usepackage{subfigure}
\usepackage{color}
\usepackage{multirow}
\usepackage{cite}
\usepackage{epstopdf}
\usepackage{pifont}
\usepackage[dvipsnames]{xcolor}

\usepackage[T1]{fontenc}    
\usepackage{url}            
\usepackage{booktabs}       
\usepackage{amsfonts,bm}       
\usepackage{nicefrac}       
\usepackage{microtype}      
\usepackage{natbib}
\usepackage{graphicx,float,subfigure}
\usepackage{lipsum}
\usepackage{caption}
\usepackage[labelfont=bf]{caption}
\usepackage{array}

\graphicspath{{figures/}}

\newcommand{\Paragraph}[1]{\vspace{-0mm} \noindent \textbf{#1} \hspace{0mm}}

\newcommand*{\boxedcolor}{red}
\makeatletter
\renewcommand{\boxed}[1]{\textcolor{\boxedcolor}{%
  \fbox{\normalcolor\m@th$\displaystyle#1$}}}
\makeatother

\makeatletter
  \newcommand\figcaption{\def\@captype{figure}\caption}
  \newcommand\tabcaption{\def\@captype{table}\caption}
\makeatother

\begin{document}
\title{Aurora Guard: Real-Time Face Anti-Spoofing via Light Reflection}

\author{
    Yao Liu$^{\dag}$ Ying Tai$^{\dag}$\thanks{\ Corresponding author} Jilin Li$^{\dag}$ Shouhong Ding$^{\dag}$ Chengjie Wang$^{\dag}$ Feiyue Huang$^{\dag}$  \\
    Dongyang Li$^{\dag}$ Wenshuai Qi$^{\ddag}$ Rongrong Ji$^{\S}$ \\
$^{\dag}$Youtu Lab, Tencent ~~~ $^{\ddag}$Shanghai University ~~~ $^{\S}$Xiamen University \\
{\tt\small $^{\dag}$\{starimeliu, yingtai, jerolinli, ericshding, jasoncjwang, garyhuang, dongyangli\}@tencent.com}\\
{\tt\small $^{\ddag}$qwspph@i.shu.edu.cn $^{\S}$rrji@xmu.edu.cn}
}

\maketitle

\begin{abstract}
   In this paper, we propose a light reflection based face anti-spoofing method named Aurora Guard (AG), which is fast, simple yet effective that has already been deployed in real-world systems serving for millions of users.
   Specifically, our method first extracts the normal cues via light reflection analysis, and then uses an end-to-end trainable multi-task Convolutional Neural Network (CNN) to not only recover subjects' depth maps to assist liveness classification, but also provide the light CAPTCHA checking mechanism in the regression branch to further improve the system reliability.
   Moreover, we further collect a large-scale dataset containing $12,000$ live and spoofing samples, which covers abundant imaging qualities and Presentation Attack Instruments (PAI).
   Extensive experiments on both public and our datasets demonstrate the superiority of our proposed method over the state of the arts.
\end{abstract}

\section{Introduction} \label{introduction}
Face anti-spoofing is currently a promising topic in computer vision research community, and also a very challenging problem in remote scenarios without specific hardware equipped in industry.
The existing methods~\cite{liu2018learning,xie2017one,yi2014face} on face anti-spoofing are paying more attention on multi-modality information ($\textit{e.g.}$, depth or infrared light).
With the development of depth sensors, recent methods and commercial systems exploit hardwares that can be embedded with structured light (\emph{e.g.}, FaceID on iphone X), light field~\cite{xie2017one} or LIDAR to reconstruct accurate $3$D shape, which can well address the limitation of $2$D methods towards high-level security~\cite{li2017face,li2016original}. Although these methods can achieve good classification performance, they highly rely on the customized hardware design, which unavoidably increases the system cost.

As a replacement, recent advances on Presentation Attack Detection~(PAD) tend to estimate depth directly from a single RGB image.
In particular, since $3$D reconstruction from a single image is a highly under-constrained task due to the lack of strong prior of object shapes, such methods introduce certain prior by recovering sparse~\cite{wang2013face} or dense~\cite{atoum2017face,liu2018learning} depth features.
However, these methods still suffer from missing the solid depth clue.
As a result, the corresponding liveness classifiers are hard to generalize to real presentation attacks in the wild.

\begin{figure}[tbp]
\centering
\setlength{\abovecaptionskip}{3mm}
\setlength{\belowcaptionskip}{-3mm}
\setlength{\lineskip}{\medskipamount}
\includegraphics[width=0.84\linewidth]{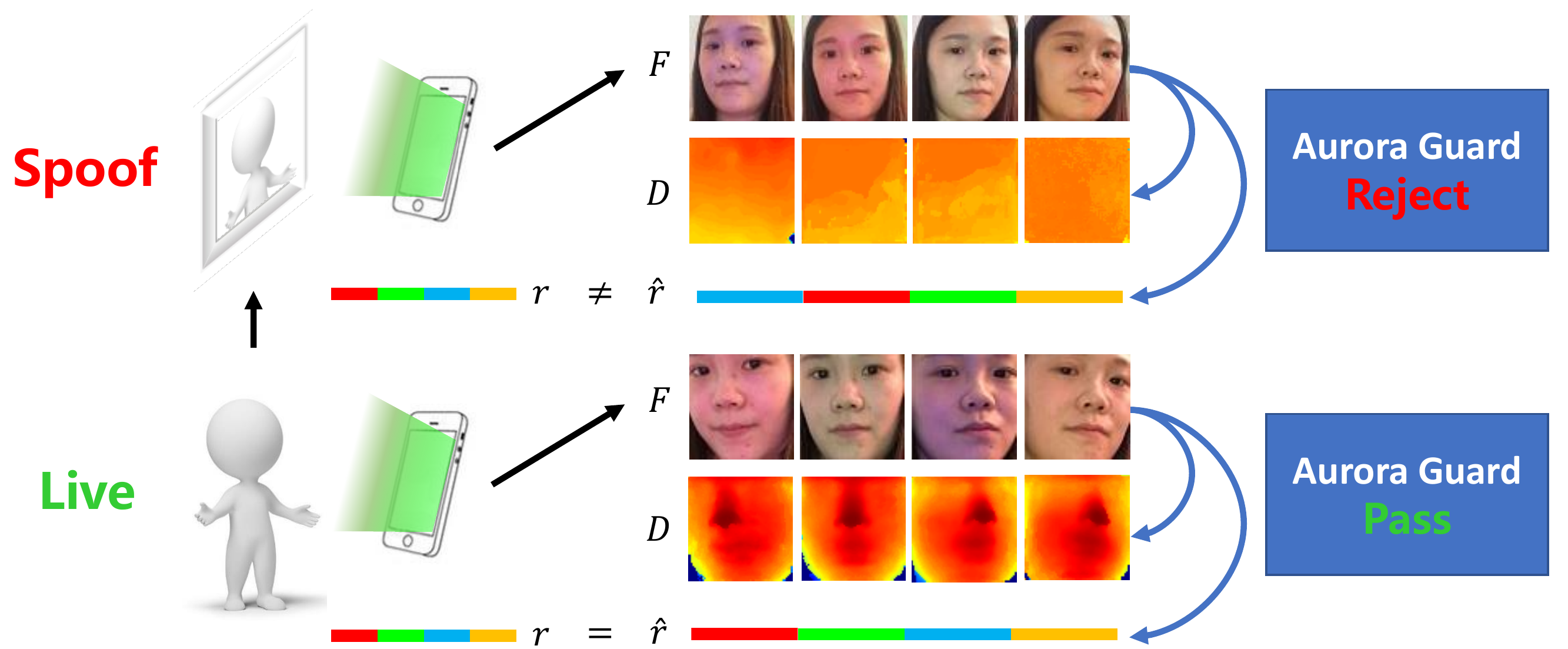}
\caption{\small \textbf{Framework of our proposed method}.
 $F_r$ denotes the facial reflection frame. $D$ denotes the recovered depth map from solid depth clue, which improves our anti-spoofing performance against unlimited $2$D spoofing.
 $r$ denotes the light CAPTCHA generated and casted by light source, and $\hat r$ is estimated by our method.
 }
\label{Fig1}
\end{figure}

Towards high accuracy and security without using depth sensors, we propose a simple, fast yet effective face anti-spoofing method termed Aurora Guard (AG).
Its principle is to use light reflection to impose two auxiliary information, \emph{i.e.}, the depth map and light parameter sequence, to improve the accuracy and security of PAD respectively (as shown in Fig.~\ref{Fig1}).
In this paper, we propose and define the light parameters sequence as \emph{light CAPTCHA}.
By only incorporating a single extra light source to generate the reflection frames, our method holds the \textit{efficiency} and \textit{portability} of cost-free software methods, which has already been deployed on smart phones and embedded terminals that serves for \textit{millions of users}.

\begin{figure*} [t!]
\centering
\setlength{\abovecaptionskip}{2mm}
\setlength{\belowcaptionskip}{-1mm}
\setlength{\lineskip}{\medskipamount}
\includegraphics[width=16.5cm]{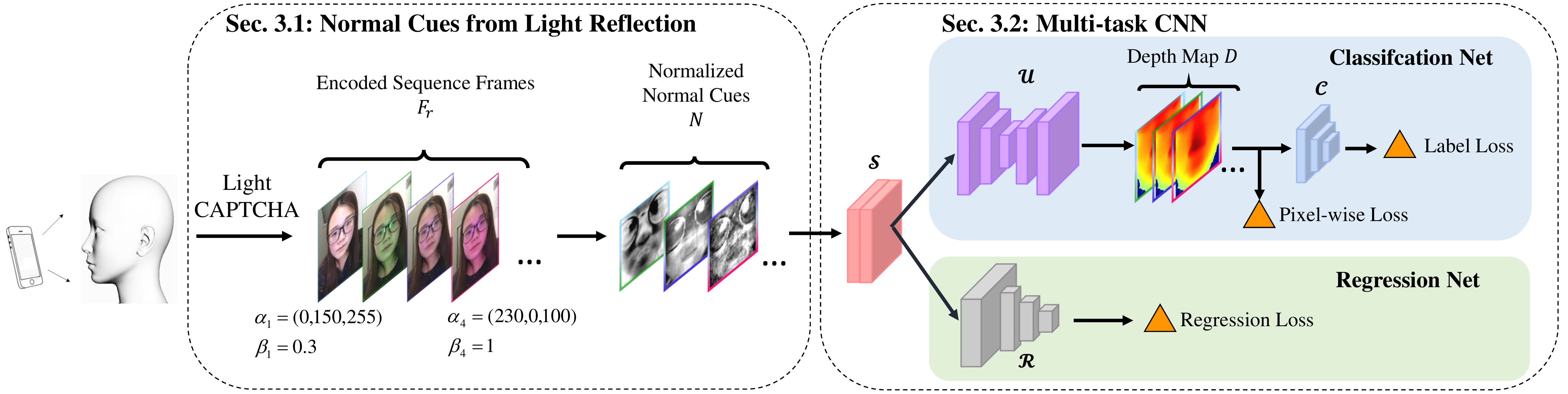}
\caption{\small \textbf{Overview of Aurora Guard}.
From facial reflection frames encoded by casted light CAPTCHA, we estimate the normal cues.
In the classification branch, we recover the depth maps from the normal cues, and then perform depth-based liveness classification. In the regression branch, we obtain the estimated light CAPTCHA.}
\label{Pipeline}
\end{figure*}

More specifically, our method mainly consists of two parts:
($1$) 
Based on Lambertian model, we cast dynamic changing light specified by the random light CAPTCHA, and then extract the normal cues from facial reflection frames.
($2$) We use an end-to-end trainable multi-task CNN to conduct liveness classification and light CAPTCHA regression simultaneously.
The classification branch estimates the depth maps from the normal cues, and classify liveness from the recovered depth map via a compact encoder-decoder structure.
The regression branch estimates the light parameter sequence, which forms a light CAPTCHA checking mechanism to handle one special type of attack named \textit{modality spoofing}, which is a very common attack in real scenarios.

Moreover,
since the imaging quality (resolution, device) and the types of Presentation Attack Instruments (PAI) are essential to evaluate performance in practical remote face authentication, we further build a dataset containing videos of facial reflection frames collected by our system, which is the most \textit{comprehensive} and \textit{largest} one in both aspects compared with other public datasets.

To sum up, the main contributions of this work include:

$\bullet$ A simple, fast yet effective face anti-spoofing method is proposed, which is practical in real scenarios \textit{without} the requirement on specific hardware design. %

$\bullet$ A \textit{cost-free} depth recover net is proposed to estimate the facial depth maps via the \textit{normal cues} extracted from the reflection frames for liveness classification.

$\bullet$ A novel \textit{light CAPTCHA checking mechanism} is proposed to significantly improve the security against the attacks, especially the modality spoofing.

$\bullet$ A dataset containing comprehensive spoof attacks on various imaging qualities and mobile ends is built.

\section{Related Work}


\Paragraph{Local Texture based Methods}
The majority of common presentation attacks are the recaptured images shown on printed photo and screens, in which the textures are different from the original images and can be leveraged to counter face spoofing.
For example, 
\cite{DBLP:journals/tifs/WenHJ15} adopted image distortion information as countermeasure against spoofing.
\cite{li2017face} proposed Deep Local Binary Pattern (LBP) to extract LBP descriptors on convolutional feature map extracted by CNN.
\cite{boulkenafet2017face} converted the face image from RGB color space to HSV-YCbCr space and extracted channel-wise SURF features~\cite{bay2006surf} to classify liveness result.
However, since the above methods operate on $2$D images, they still suffer from poor generalization ability to unseen attacks and complex lighting conditions, especially when RGB sensors have low resolution or quality.
In contrast, our method exploits $3$D information (\emph{e.g.}, depth) via the reflection increments from RGB images, which makes our method more robust and accurate to various attacks.

\Paragraph{Depth Sensor based Methods}
It is well known that the $3$D facial cues can be used to defeat $2$D presentation attacks.
For example,~\cite{wang2017robust} directly exploit depth sensors such as Kinect to recover depth map and evaluate the anti-spoofing effectiveness combined with texture features.
\cite{xie2017one} introduced a light field camera to extract depth information from multiple refocused images took in one snapshot.
Moreover, iPhone X incorporates a structured-light sensor to recover accurate facial depth map, which obtains impressive anti-spoofing performance.
However, although iPhone X achieves high accuracy, there are two practical problems.
First, it uses an \textit{expensive} $3$D camera for accurate depth.
Second, its implementation details are missing.
In contrast, our method is not only hardware-free that has competitive results against $3$D hardware via a \textit{cost-free} depth recover net, but also easy to follow for \textit{re-implementation}.



\Paragraph{Depth Estimated from Single Image}
\cite{wang2013face} firstly attempted to recover a sparse $3$D facial structure from RGB image for face anti-spoofing.
\cite{atoum2017face} proposed a two-steam depth-based CNN to estimate both texture and depth.
Recently,~\cite{liu2018learning} fused multiple sequential depth predictions to regress to a temporal rPPG signal for liveness classification.
However, $3$D reconstruction from a single image is still highly under-constrained, since these methods suffer from missing solid $3$D information clue.
As a result, their anti-spoofing classifiers are hard to generalize to unseen spoof attacks, and is also sensitive against the quality of RGB camera.
To address the inaccurate depth issue, our method first obtains normal cues based on the light reflection, which better removes the effects of albedo and illuminance.
Then we train a compact encoder-decoder network to accurately recover the depth map. 

\Paragraph{Lambertian Reflection based Methods}
\cite{tan2010face} firstly pointed out the importance of Lambertian modeling for face anti-spoofing, while only obtains rough approximations of illuminance and reflectance parts.
\cite{chan2018face} also adopted Lambertian reflection model to extract simple statistics (\emph{i.e.}, standard deviation and mean) as features, but achieves limited performance.
Our method differs from the above methods in three key aspects:
($1$) We \textit{actively} perform light reflection via an extra light source specified by random light parameter sequence, while the above methods do NOT.
($2$) We construct a regression branch to achieve the novel \textit{light CAPTCHA} checking mechanism to make the system more robust, while the above methods again lack such scheme.
($3$) We incorporate deep networks to learn powerful features, while the above methods use simple handcrafted features.


\section{The Proposed Method}


Fig.~\ref{Pipeline} illustrates the entire process of our method.
Specifically, we first set a smart phone (or any other devices) with front camera and light source ($\textit{e.g.}$, the screen)
in front of the subject.
Then, a random parameter sequence ($\textit{i.e.}$, light CAPTCHA) of light hues and intensities is generated, \emph{i.e.}, $r=\{(\alpha_i, \beta_i)\}_{i=1}^{n}$ given $n$ frames.
We manipulate the screen to cast dynamic light specified by the light CAPTCHA $r$.
After the reflection frames $F_r$ are captured, we sequentially estimate the normal cues $N$, 
which are the input of a multi-task CNN to predict liveness label and regress the estimated light CAPTCHA $\hat r$. The final judgement is been made from both of the predicted label and the matching result between $\hat r$ and $r$.

%

\begin{figure}[t!]
\centering
\setlength{\abovecaptionskip}{2mm}
\setlength{\belowcaptionskip}{-3mm}
\setlength{\lineskip}{\medskipamount}
\includegraphics[width=0.9\linewidth]{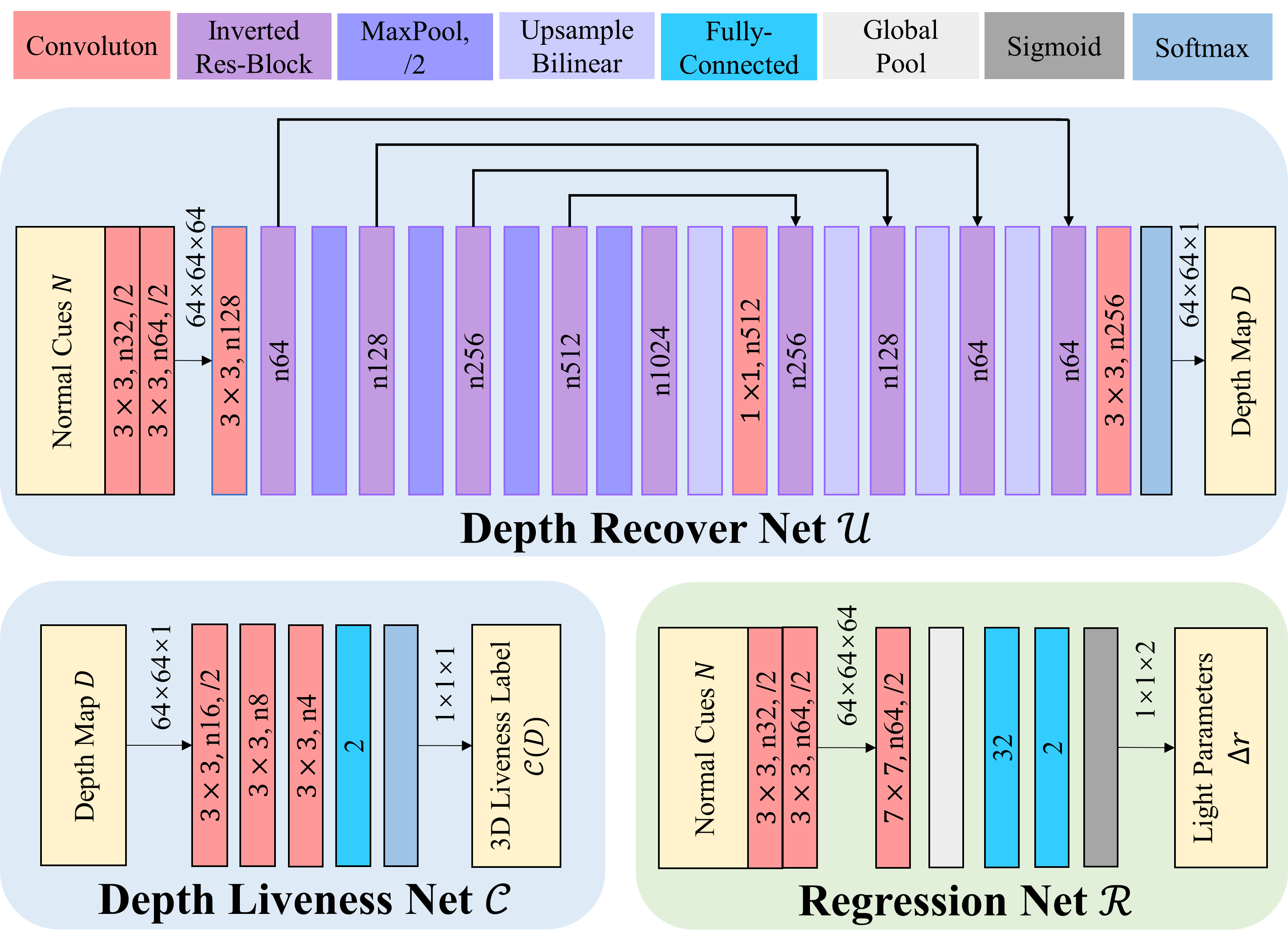}
\caption{\small \textbf{The architecture details of the proposed multi-task CNN}. Here $n$ denotes the number of output feature maps.}
\label{DepthMapNet}
\end{figure}

\subsection{Normal Cues from Light Reflection}
\label{3.1}

Given the facial reflection frames $\{F_{r_i}\}_{i=1}^n$,
we extract normal cues by estimating the reflection increments on subject's face.
Since objects with rough surfaces are diffuse reflectors (\emph{e.g.} human face), light casted onto surface point is scattered and reflected, and then perceived as the final imaging in the camera.
Given images containing reflection on the object surface, we measure the magnitude variations among different images, under the assumption of Lambertian reflection model and weak perspective camera projection.

Lambert's Law regards the reflected part to be equal on all directions on the diffuse surface.
In other words, for any pixel point $\mathbf{p}$ of the camera image under specific casting light $L_{r}$, its intensity $F_{r}(\mathbf{p})$ is formulated as:
\begin{equation}
F_{r}(\mathbf{p}) = \rho_{p}(k_{a}+k_{r}\mathbf{l}\cdot \mathbf{n}_{p}),
\end{equation}
where $k_{a}$ is the ambient weight, $k_{r}$ is the diffuse weight, $\mathbf{l}$ is the light source direction, $\rho_{p}$ is the albedo and $\mathbf{n}_{p}$ is the point normal.
When light changes {suddenly}, $k_{a}$ and $\mathbf{l}$ (position of the screen) are not supposed to change temporally and can be regarded as constants. We adopt affine transformation to align $\mathbf{p'}$ and $\mathbf{p}$ between image pairs, with transformation matrix estimated from the facial landmarks detected by PRNet~\cite{feng2018prn}. Then in another image under casting light $L_{r'}$, the intensity of the registered pixel $\mathbf{p'}$ is:

\begin{equation}
F_{r'}(\mathbf{p}) = F_{r'}(\mathbf{p'}) = \rho_{p'}(k_{a}+k_{r'}\mathbf{l}\cdot \mathbf{n}_{p'}).
\end{equation}
We then attain the scalar product $N_{\Delta r}(\mathbf{p})$ on each point, 
\begin{equation}
N_{\Delta r}(\mathbf{p}) = \mathbf{l}\cdot\mathbf{n}_p = \frac{F_{r}(\mathbf{p}) - F_{r'}(\mathbf{p})}{k_{r} - k_{r'}},
\end{equation}
where the scalar map arranged by $N_{\Delta r}(\mathbf{p})$ is the \textit{normal cue}. 

\subsection{Multi-task CNN}

After obtaining the normal cues, we adopt a multi-task CNN that has two branches to achieve liveness classification and light CAPTCHA regression, respectively.
It should be noted that our multi-task structure is \textit{task-driven}, which enables double checking mechanism to improve the robustness on modality spoofing in practical scenarios.

\Paragraph{Liveness Classification.}
Depending on the lighting environment, the normal cues extracted from facial reflection frames may be rough and noisy.
To efficiently obtain accurate depth information from the normal cues, we adopt an encoder-decoder network, which balances the performance and speed.
The network architecture is inspired by~\cite{ronneberger2015u,tai2018fsrnet}, in which we use the inverted residual block~\cite{sandler2018mobilenetv2}.
The recovered depth map is then sent to a simple classification structure to distinguish the real $3$D face from those $2$D presentation attacks.
The detailed structure is shown in Fig.~\ref{DepthMapNet}.

After obtaining $m$ frames of normal cues $N_1, N_2,...,N_m$ of one video, the classifier has the following loss function:

\vspace{-2mm}
\begin{equation}\
\small
\begin{aligned}
\mathcal{L}_{cls} = & \frac{1}{m} \sum_{i=1}^{m} \Big \{ -(1-\lambda_{depth})(c_ilog(\mathcal{C}(\mathcal{U}(\mathcal{S}(N_i)))) \\
          & + (1-c_i)log(1-\mathcal{C}(\mathcal{U}(\mathcal{S}(N_i))))) \\
          & + \lambda_{depth}\sum_{\mathbf{p} \in \mathbb{Z}^2} - k log
          (
          e^{d_k(\mathbf{p})} /
            (
            \begin{matrix}
            \sum_{k'=1}^{256} e^{d_{k'}(\mathbf{p})}
            \end{matrix}
            )
          ) \Big \},
\end{aligned}
\label{eq:ClsDepthLoss}
\end{equation}
where $\mathcal{S}$ denotes stem operation that contains two convolutional layers, $\mathcal{C}$ denotes the depth liveness prediction net, $\mathcal{U}$ denotes the depth recover net, $c_i$ is the liveness label of the $i$-th normal cue, $\lambda_{depth}$ is the weight of the depth estimation loss. In depth recovering part, we adopt $2$D pixel-wise soft-max over the predicted depth map combined with the cross-entropy loss function, where $k: \Omega \to {1, ..., 256}$ is the ground truth depth label, $d_k(\mathbf{p})$ is the feature map activation on channel $k$ at the pixel position $\mathbf{p}$, while the feature map activation $\mathcal{U}(N_i)$ is the output of the depth recover net.


\vspace{-0.5mm}
\paragraph{Light Parameter Regression.}
\label{Regression}
 We reinforce the security of our method against \textit{modality spoofing}, which is further discussed in Sec.~\ref{CAPTCHA}, by customizing the casted light CAPTCHA and exploit a regression branch to decode it back for double checking automatically.

By feeding the same normal cues as the classification branch, the regression net has the loss function $L_{reg}$ as:
\begin{equation}\
\small
\begin{aligned}
\mathcal{L}_{reg} = & \frac{1}{m} \sum_{i=1}^{m} \{ \|\mathcal{R}(\mathcal{S}(N_i)) - \Delta r_i\|^{2} \},
\end{aligned}
\end{equation}
where $\mathcal{R}$ denotes the regression net, $\Delta r_i$ is the ground truth light parameter residual of reflection frames $F_{r_{i}}$ and $F_{r_{i-1}}$.

Supposing there are $V$ videos in the training set, the entire loss function of our multi-task CNN is formulated as:
\begin{equation}\
\small
\begin{aligned}
\mathcal{L}{(\Theta)} = \mathop{\arg\min}\limits_{\mathbf{\Theta}} \frac{1}{2V} \sum_{v=1}^{V} \{ \mathcal{L}_{cls}^v + \lambda_{reg}\mathcal{L}_{reg}^v \},
\end{aligned}
\end{equation}
where $\Theta$ denotes the parameter set, $\lambda_{reg}$ is the weight of CAPTCHA regression loss.
In practice, we set the light CAPTCHA sequence to be composed by $4$ types of light in random order, which balances the robustness of CAPTCHA checking and time complexity. 
We set the rate of light changing identical to the frame rate, thus the frames hold different light reflection.
The length of $F, r$ equals to $m+1$.
The Signal-to-Noise Ratio ($SNR$) is adopted to check if the estimated light parameter sequence matches the ground truth sequence.

\subsection{Dataset Collection}
\label{Dataset}
\begin{figure}[t!]
\centering
\setlength{\lineskip}{\medskipamount}
\setlength{\abovecaptionskip}{1mm}
\setlength{\belowcaptionskip}{-3mm}
\includegraphics[width=0.96\linewidth]{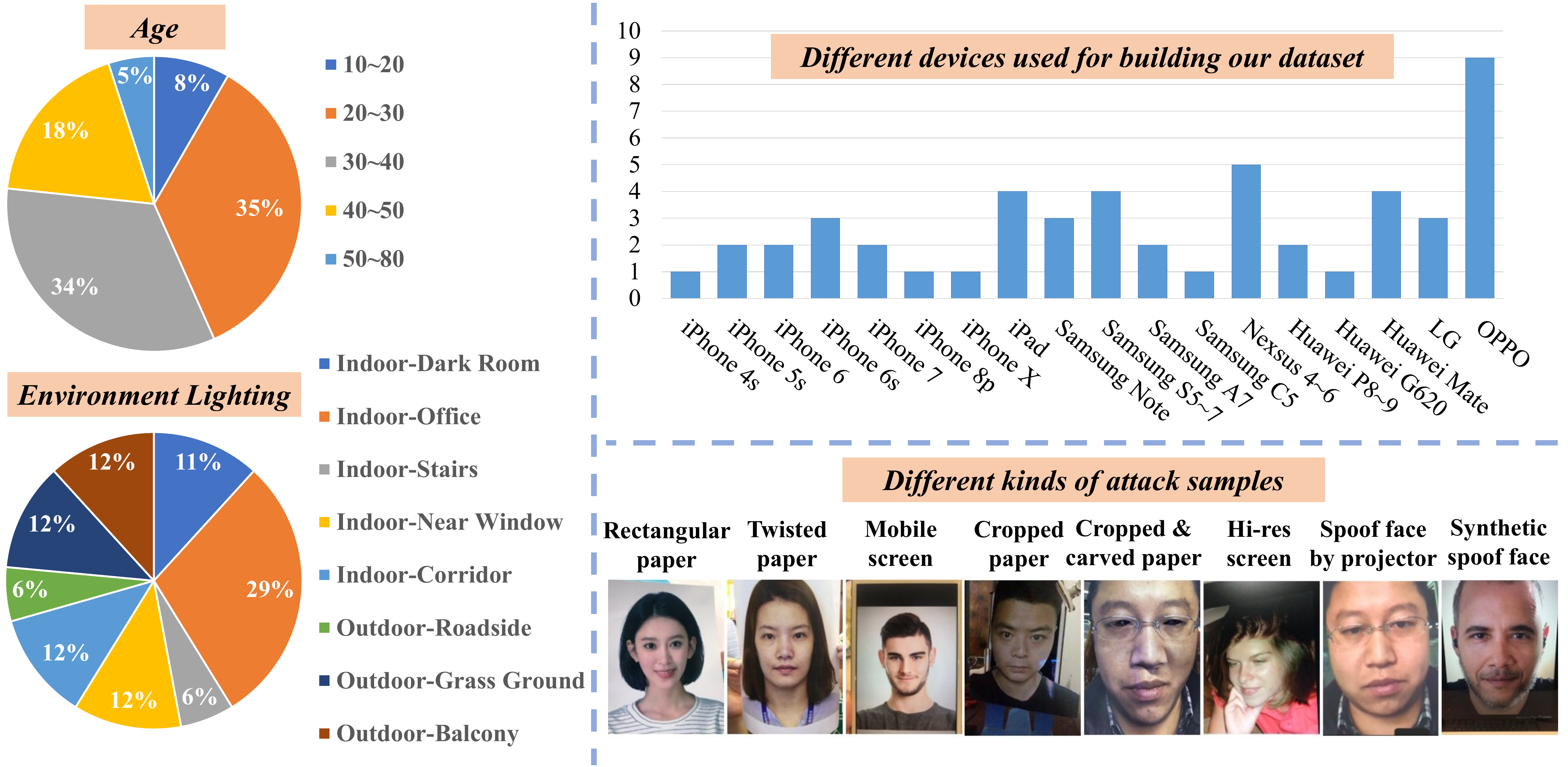}
\caption{\small \textbf{Statistics and attack samples} of our dataset.}
\label{AttackSamples}
\end{figure}

As claimed in Sec.~\ref{introduction}, various imaging qualities and the types of PAIs are very important for practical remote face authentication.
To address this need, we collect a new dataset, in which each data sample is obtained by casting dynamic light sequence onto the subject, and then record the $30$-fps videos. 
Some statistics of the subjects are shown in Fig.~\ref{AttackSamples}.
Note that we mainly collect $2$D attacks, the main target in most prior anti-spoofing methods~\cite{atoum2017face,liu2018learning}, rather than $3$D ones because the cost to produce and conduct $3$D attacks in real scenarios is much higher than $2$D attacks.

Compared to the previous public datasets~\cite{liu2018learning,DBLP:conf/biosig/ChingovskaAM12,casia2012face}, our dataset has three advantages:
$1$) Our dataset is the \textit{largest} one that includes $12,000$ live and spoof videos, with average duration to be $3$s, collected from $200$ subjects, compared to $4,620$ videos from $165$ subjects in~\cite{liu2018learning}.
$2$) Our dataset uses the \textit{most extensive} devices (\textit{i.e.}, $50$ in ours vs. $4$ in~\cite{liu2018learning}) to obtain good simulation of real-world mobile verification scenarios.
$3$) Our dataset contains the \textit{most comprehensive} attacks that include various print, replay, modality and another spoof face by light projector (see Fig.~\ref{AttackSamples}). 

We divide samples into $3$ parts through the spoof types: paper attack, screen attack and other complex attacks consisting of cropped paper photos, projection attacks, \emph{etc}.
In each part, the data is further divided into train set, develop set and test set, as shown in Tab.~\ref{protocol}.
Moreover, the amounts of live data and spoof data stay equal in our dataset.
The live data is collected under multiple variations including interference illumination on face, noisy imaging quality and different poses. 
The spoof data are collected through abundant PAIs. 

\section{Experiments}


\subsection{Implementation Details}

\vspace{-0.5mm}
\paragraph{Model Training}
We use Pytorch to implement our method and initialize all convolutional and fully-connected layers with normal weight distribution~\cite{kaimingnorm}.
For optimization solver, we adopt RMSprop~\cite{RMSprop} during training process.
Training our network roughly takes $5$ hours using a single NVIDIA Tesla P$100$ GPU and iterates for $\thicksim$$300$ epochs. 

\paragraph{Evaluation Criteria}
We use common criteria to evaluate the anti-spoofing performance, including False Rejection Rate ($FRR$), False Acceptance Rate ($FAR$) and Half Total Error Rate ($HTER$), which depends on the threshold value $\tau_{cls}$.
To be specific, $FRR$ and $FAR$ are monotonic increasing and decreasing functions of $\tau_{cls}$, respectively.
A more strict classification criterion corresponds to a larger threshold of $\tau_{cls}$, which means spoof faces are less likely to be misclassified.
For certain data set $\mathbb{T}$ and $\tau_{cls}$, $HTER$ is defined by
\vspace{-1.5mm}
\begin{equation}
\small
HTER(\tau_{cls},\mathbb{T})=\frac{FRR(\tau_{cls},\mathbb{T})+FAR(\tau_{cls},\mathbb{T})}{2}\in(0,1).
\end{equation}
Lower $HTER$ means better average performance on liveness classification, and $HTER$ reaches its minimum when $FAR$=$FRR$, which is defined as Equal Error Rate ($EER$).


\begin{table} [t!]
\centering
\setlength{\abovecaptionskip}{2mm}
\caption{\small Generalization experiment protocol. The train, development and test set are divided into $3$:$1$:$1$ within the samples in each part.}
\label{protocol}
\vspace{-2mm}
\scalebox{0.87}{
    \begin{tabular}{| c | c | c | c |}
    \hline
    Part                                                    & Type                & Samples                & Collection        \\ \hline
    \multirow{2}{*}{Part~1}    & Paper Attack          & 2000               & Phone No.~1$\sim$17           \\ \cline{2-4}
                                                                      & Live Person            & 2000               & Subject No.1$\sim$70              \\ \hline
    \multirow{2}{*}{Part~2}    & Screen Attack        & 2200               & Phone No.~18$\sim$34         \\ \cline{2-4}
                                                                      & Live Person            & 2000               & Subject No.71$\sim$140         \\ \hline
    \multirow{2}{*}{Part~3}    & Complex Attack    & 1800               & Phone No.~35$\sim$50       \\ \cline{2-4}
                                                                      & Live Person            & 2000               & Subject No.141$\sim$200     \\ \hline
    \end{tabular}
    }
\end{table}


\begin{table} [t!]
\small
\centering
\setlength{\abovecaptionskip}{2mm}
\caption{\small Comparisons of $EER$ from development set and $HTER$ from testing set in our dataset.}
\label{DepthLossExp}
\vspace{-2mm}
\scalebox{0.85}{
    \begin{tabular}{|c|c|c|c|c|}
    \hline
     $\lambda_{depth}$    &  $0.0$     & $0.2$   & $0.4$      & $0.5$      \\ \hline
     EER~(\%)             & $4.79\pm0.41$   & $2.31\pm0.23$    & $1.58\pm\textbf{0.19}$ & $\textbf{1.48}\pm0.21$          \\ \hline
     HTER~(\%)            & $7.20\pm0.77$      & $3.53\pm0.43$  & $2.21\pm\textbf{0.31}$    & $\textbf{2.09}\pm0.33$       \\ \hline
    \end{tabular}
    }
\end{table}

\begin{table} [t!]
\centering
\setlength{\abovecaptionskip}{2mm}
\caption{\small Quantitative evaluation on modality spoofing.}
\label{RandomAttack}
\vspace{-2mm}
\scalebox{0.8}{
    \begin{tabular}{| c | c | c | c | c | c |}
    \hline
     & Video 1 & Video 2 & Video 3 & Video 4 & Video 5  \\ \hline
    \multirow{2}{*}{FAR} & 2/3000 & 0/3000 & 0/3000 & 1/3000 & 0/3000 \\ \cline{2-6}
     & 0.06\% & 0.00\% & 0.00\% & 0.03\% & 0.00\% \\ \hline
    \end{tabular}
    }
\end{table}

\vspace{-1.5mm}
\subsection{Ablation Study}
\paragraph{Effectiveness of Depth Supervision}
First, we conduct experiments to demonstrate the effects of the depth supervision.
To be specific, we monotonically increase the weight of depth loss $\lambda_{depth}$ in Eq.~\ref{eq:ClsDepthLoss} and train multiple models, respectively.
Under each $\lambda_{depth}$, we train $10$ different models, and then evaluate the mean and standard variance of $EER$ and $HTER$, as shown in Tab.~\ref{DepthLossExp}.
When $\lambda_{depth}$=$0$, the normal cues are directly used for liveness classification, which achieves the worst results.
As we increase the $\lambda_{depth}$ to give more importance on the auxiliary depth supervision, the performance improves gradually, which verifies its effectiveness to helps denoise the normal cues and enhance the $3$D information.

\begin{figure}[t!]
\centering
\setlength{\lineskip}{\medskipamount}
\setlength{\abovecaptionskip}{2mm}
\setlength{\belowcaptionskip}{-4mm}
\includegraphics[width=0.9\linewidth]{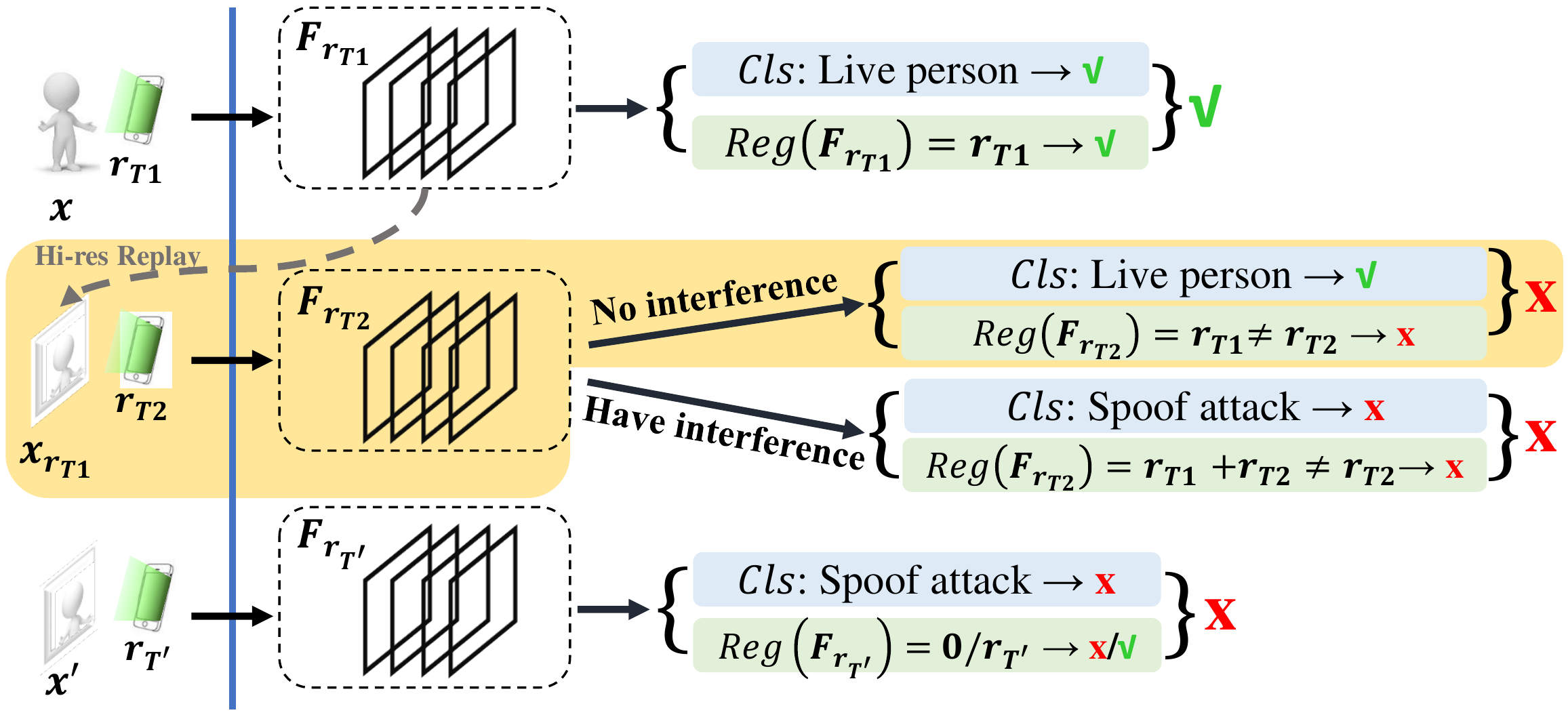}
\caption{\small \textbf{Illustration on our double checking mechanism}.
$Cls, Reg$ are the classification net and regression net, respectively.
$1$) The first row handles live person.
$2$) The highlighted yellow part in the second row represents \textit{modality spoofing} (\textit{i.e.}, $x_{r_{T1}}$), which replays the formerly captured Hi-res video frames $F_{r_{T1}}$ that contains true facial reflection, which fools the $Cls$ but can be defended by the light CAPTCHA checking scheme in $Reg$.
$3$) No interference indicates the reflection effect caused by $r_{T_2}$ is \textbf{blocked}, thus $F_{r_{T2}}$ shares similar facial reflection with $F_{r_{T1}}$ and can pass the $Cls$.
$4$) The bottom row indicates the conventional $2$D spoofing case.}
\label{ModalitySpoofing}
\end{figure}

\begin{figure}[t!]
\centering
\setlength{\lineskip}{\medskipamount}
\setlength{\abovecaptionskip}{2mm}
\setlength{\belowcaptionskip}{-2mm}
\includegraphics[width=1.0\linewidth]{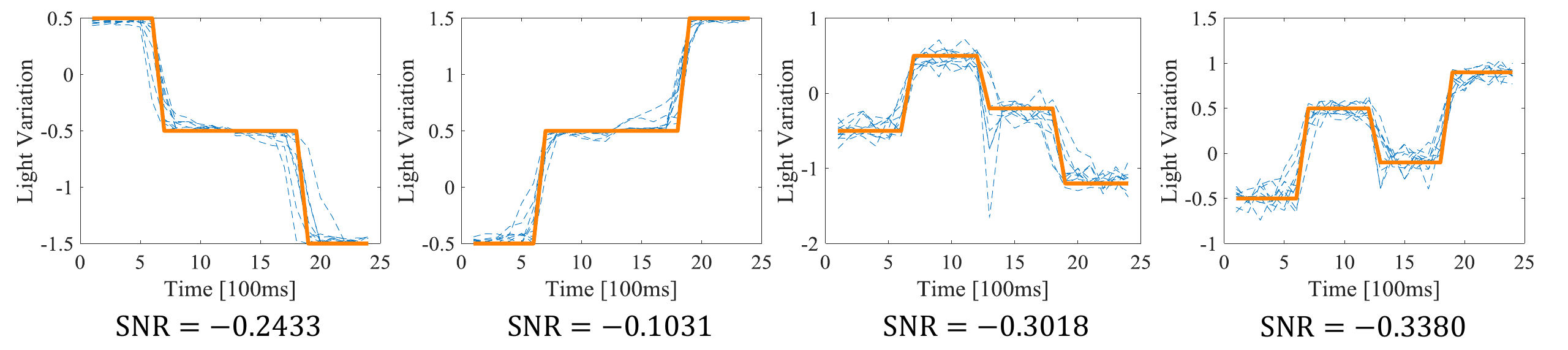}
\vspace{-5mm}
\caption{\small
\textbf{Illustration on estimated light CAPTCHA}.
Each figure shows $10$ estimated curves obtained by our regression branch (blue dotted) from different subjects and scenes compared to the ground truth (orange solid), where the x-axis and y-axis denote the time and temporal variation of light hue $\alpha$ respectively. }
\label{LightRegressionExp}
\end{figure}

\paragraph{Light CAPTCHA Regression Branch}\label{CAPTCHA}
Although our system can well handle most of the normal $2$D presentation attacks via depth information, it may still suffer from one special spoofing attack named \textit{modality spoofing}, which directly forges the desired reflection patterns.
Specifically, modality spoofing will fail our classification net when meeting $2$ requirement:
$1$) The formerly captured raw video consists of facial reflection frames, that contains the true reflection patterns, is leaked and replayed by Hi-res screen.
$2$) Within capture process of attack trial, the casted light doesn't interfere with the original facial reflection in video frames. 
Fig.~\ref{ModalitySpoofing} illustrates the principle of our light CAPTCHA against the modality spoofing.
We further conduct experiments to prove the effectiveness of our light CAPTCHA checking mechanism in Fig.~\ref{LightRegressionExp}.
The $|SNR|$ results of our regression branch are all below $0.35$ and close with the ground truth CAPTHCHA, which demonstrates its ability to distinguish $4$ types of casting light. 

Next, we quantitatively verify the effectiveness of our method against the modality spoofing attacks.
We set the light CAPTCHA formed as the compositions of $4$ types of casting light in a random order for every checking trial.
To perform modality spoofing, we record videos consisting the same $4$ types of true facial reflection (\textit{i.e.}, $F_{r_{T1}}$ in Fig.~\ref{ModalitySpoofing}), and repeatedly replay the video for $3,000$ times. 
In other words, the fixed video loop must match the randomly generated CAPTHCHA to bypass our system. 
The experiment results in Tab.~\ref{RandomAttack} show that the light CAPTCHA checking mechanism highly improves the security on modality spoofing. 

\begin{figure}
\centering
\setlength{\lineskip}{\medskipamount}
\setlength{\abovecaptionskip}{2mm}
\setlength{\belowcaptionskip}{-2mm}
\includegraphics[width=0.95\linewidth]{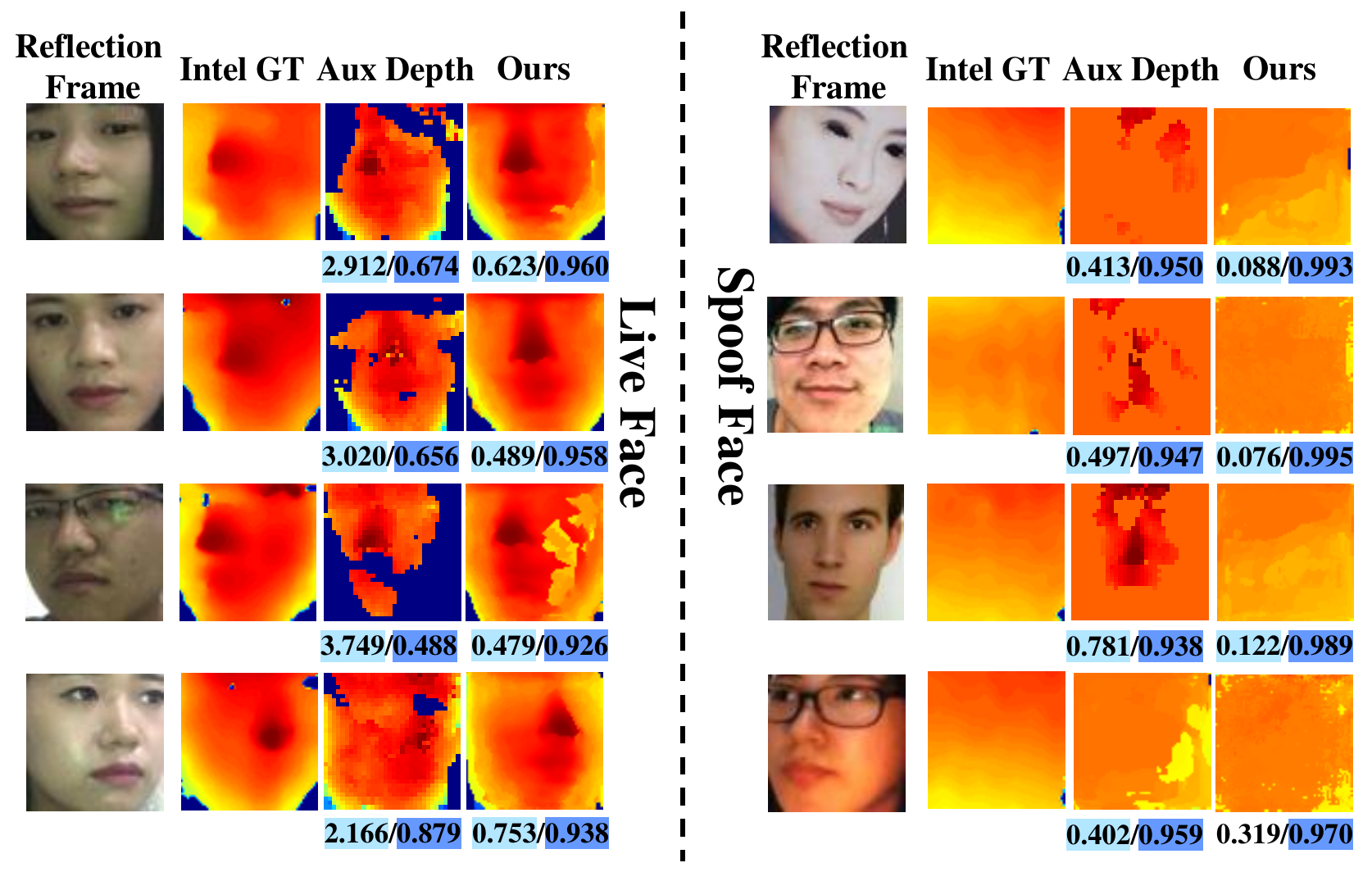}
\vspace{-2mm}
\caption{\small
\textbf{Comparisons on depth recovery}.
We take the depth data from Intel $3$D camera as the ground truth.
Results are computed using the depth metrics from~\cite{godard2017unsupervised}.
The light blue $RMSE(log)$ measures error in depth values from the ground truth (Lower is better).
And the dark blue $\delta < 1.25$ measures error in the percentage of depths that are within threshold from the correct value (Higher is better).
Note that Aux Depth~\cite{liu2018learning} recovers depth map from single RGB image, while ours recovers from reflection frames which contain solid depth clues.
The better recovered depth enables our method to accurately classify liveness, without additional texture or rPPG supervision.}
\label{DepthMapRecover}
\end{figure}

\begin{table}[t!]
\small
\centering
\setlength{\abovecaptionskip}{2mm}
\caption{\small Comparison of $EER$ from development set and $HTER$ from test set in our data set.}
\label{HTER}
\vspace{-2mm}
\scalebox{0.93}{
    \begin{tabular}{|c|c|c|c|c|}
    \hline
     Method       & EER~(\%)     & HTER~(\%)     \\ \hline
     SURF~\cite{boulkenafet2017face}     & $4.72$     & $14.65$                                         \\ \hline
     Deep LBP~\cite{li2017face}     & $5.61$     & $8.83$                                            \\ \hline
     FASNet~\cite{FASNet}     & $5.67$       &  $8.60$                                            \\ \hline
     Auxiliary Depth CNN~\cite{liu2018learning}     & $2.55$      & $5.36$                \\ \hline
     Ours      & $\textbf{1.24}$   & $\textbf{1.91}$                          \\ \hline
    \end{tabular}
    }
\end{table}

\subsection{Comparison to State-of-the-Art Methods}\label{4.4}

\paragraph{Depth Map Estimation}
Next, we conduct comparisons on depth recovery against the recent state-of-the-art method~\cite{liu2018learning}, as shown in Fig.~\ref{DepthMapRecover}.
We see that our method can recover more accurate depth map on various aspects, such as pose, facial contour and organ details, which demonstrate the effects to recover depth from solid depth clue instead of RGB texture.
It should also be noted that our method achieves comparable results to the Intel $3$D sensor that can absolutely detect $2$D presentation attacks without failure cases.

\paragraph{Face Anti-Spoofing}
Here, we conduct comparisons on anti-spoofing, in which our method and several state-of-the-art methods are trained on our dataset (\textit{i.e.}, all the $3$ training sets in each part), and then tested on public and our datasets, respectively.
After training, we determine the threshold $\tau_{cls}$ via the $EER$ on the develop set and evaluate the $HTER$ on the test set.
First, we conduct test on our dataset. 
Tab.~\ref{HTER} shows that our method significantly outperforms the prior methods, where Aux Depth~\cite{liu2018learning} ranks $2$nd, while the conventional texture based methods~\cite{boulkenafet2017face,li2017face} achieve relatively lower performance.

Next, we conduct tests on two public datasets: Replay-Attack~\cite{DBLP:conf/biosig/ChingovskaAM12} and CASIA~\cite{casia2012face}.
To better show the effectiveness and generalization of our method, \emph{NO} additional fine-tuning is performed.
Since our method requires casting extra light onto the subjects, the only way to test the live subjects is to let the real person involved in the public dataset to be presented, which is impossible and unable us to measure $FRR$ on public dataset.
For the spoof samples in these two public datasets, we print or broadcast the videos to act as the negative subjects and evaluate the $FAR$ of various methods in Tab.~\ref{FAR on public set}.
The results again demonstrate the effectiveness and generalization of our method compared to the state-of-the-art methods.

\begin{figure}[t!]
\centering
\setlength{\lineskip}{\medskipamount}
\setlength{\abovecaptionskip}{2mm}
\setlength{\belowcaptionskip}{-1mm}
\includegraphics[width=1.0 \linewidth]{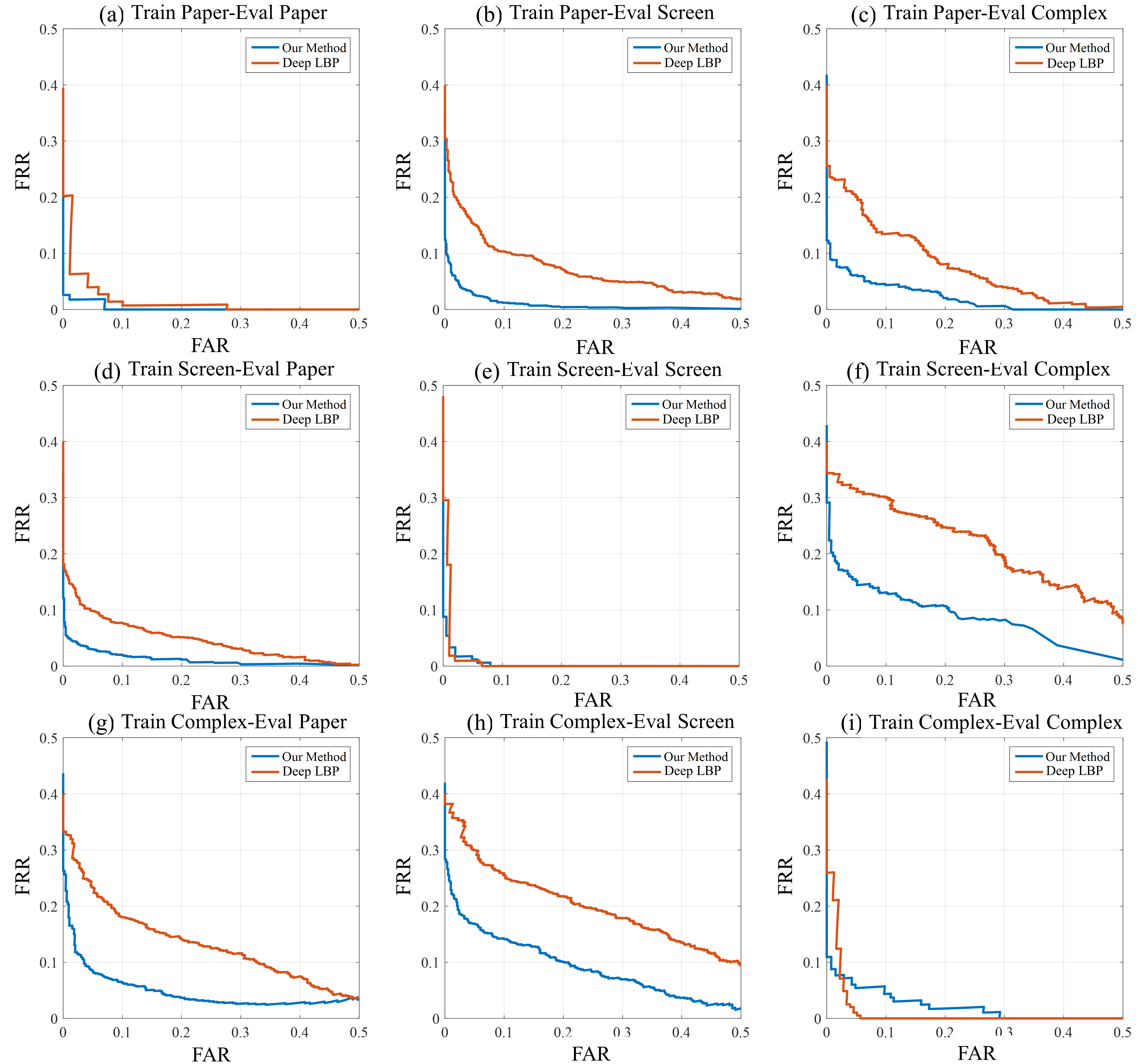}
\caption{\small \textbf{Generalization experiments} with training and testing pairwisely on every pair of sub-dataset combination.}
\label{fig:5}
\end{figure}

\paragraph{Model Generalization} 
Robust generalization ability is a key characteristic for face anti-spoofing in real scenarios. 
Here, we conduct generalization comparisons with a state-of-the-art local texture based method~\cite{li2017face}, including live subject identity, device sensor interoperability and types of presentation attack.
Specifically, we train both models from training set of each part in our dataset, and evaluate the performance in all three test sets respectively.
Each model only learns one type of spoof attack with partial subject identities as well as device sensors, and the results are shown in Fig.~\ref{fig:5}. 
We can conclude that:
$1$) Curves (a),(e),(i) show that both models can achieve ideal $FRR$ and $FAR$ when training and evaluating are performed on the same attack.
$2$) Curves (b),(d) show that when testing on unseen attacks, the $EER$ of~\cite{li2017face} degrades to~$\thicksim$$0.10$ while our method is still below $0.04$ with degradation of only~$\thicksim$$0.01$.
$3$) Harder negatives from complex dataset lead to worse performance, where the $EER$ of~\cite{li2017face} in {(i)} rises from $0.03$ to $0.22$ in {(h)}, while our method only goes up to $0.13$.
$4$) In the other cases, our method retains half of the degradation on $FAR$ and $FRR$, which demonstrates its robust generalization ability.

\begin{table} [t!]
\small
\centering
\setlength{\abovecaptionskip}{2mm}
\caption{\small $FAR$ indicator cross-tested on public dataset. Here to mention we use the same model trained from our data set without finetuning and same $\tau_{cls}$ to evaluate $FAR$ on public dataset.}
\label{FAR on public set}
\vspace{-2mm}
\scalebox{0.9}{
    \begin{tabular}{|c|c|c|c|c|c|c|}
    \hline
    \multirow{2}{*}{Method}    & Replay-Attack        & CASIA                 \\ \cline{2-3}
     & FAR(\%) & FAR(\%) \\ \hline
    Color texture~\cite{boulkenafet2015face}    & $0.40$                            & $6.20$                     \\ \hline
    Fine-tuned VGG-face~\cite{li2016original}    & $8.40$                            & $5.20$                     \\ \hline
    DPCNN~\cite{li2016original}    & $2.90$                            & $4.50$                      \\ \hline
    SURF~\cite{boulkenafet2017face}    & $0.10$                             & $2.80$                      \\ \hline
    Deep LBP~\cite{li2017face}    & $0.10$          & $2.30$                      \\ \hline
    Patch-Depth CNNs~\cite{atoum2017face}    & $0.79$                              & $2.67$                      \\ \hline
    Ours     & $\textbf{0.02}$          & $\textbf{0.75}$    \\ \hline
    \end{tabular}
    }
\end{table}

\begin{figure}[t!]
\centering
\setlength{\lineskip}{\medskipamount}
\setlength{\abovecaptionskip}{2mm}
\setlength{\belowcaptionskip}{-2mm}
\includegraphics[width=0.88\linewidth]{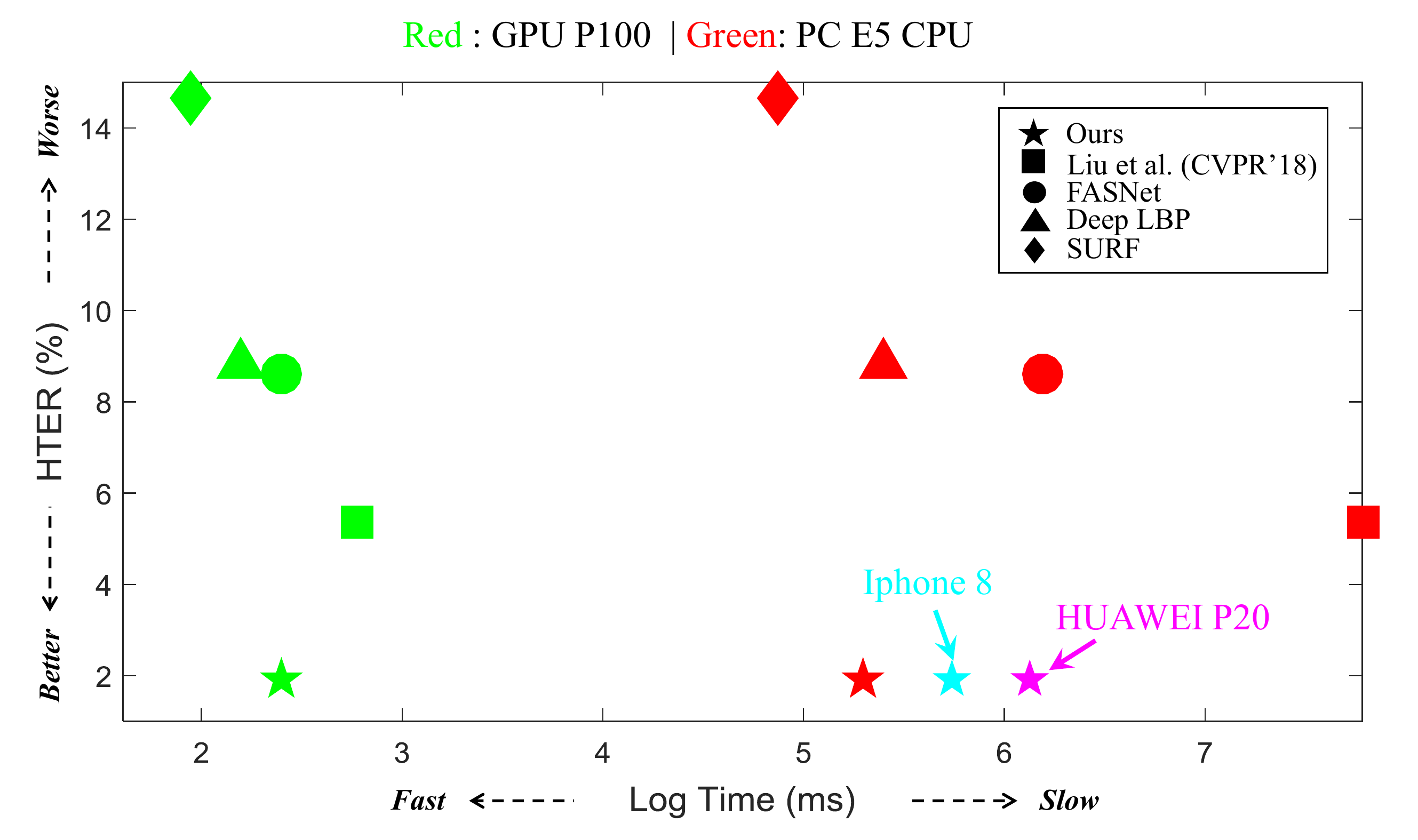}
\caption{\small\textbf{ Time Comparison} between several SOTA methods and ours in the aspects of effectiveness and cross-platform efficiency.}
\label{TimeComplexity}
\end{figure}

\vspace{-0.5mm}
\paragraph{Running Time Comparison}
We compare cross-platform inference time with several SOTA methods.
We deploy and compare on $3$ common platform architectures: GPU for cloud server, CPU (x$86$) for some embedded chips and CPU (arm) for smart phones, as shown in Fig.~\ref{TimeComplexity}.
As we can see, our efficiency on mobile platform still meets the application requirement and even outperforms some methods on CPU (x$86$).
The results indicate that our method achieves real-time efficiency and is portable for cross-platform computation requirements along with state-of-the-art anti-spoofing performance.

\subsection{Comparison to Hardware-based Method}
\label{Compare3D}

Finally, we compare our method with Structured-Light $3$D~(SL$3$D), which relies on the hardware that is embedded with structured-light $3$D reconstruction algorithm. 
To be specific, we use an Intel$^\circledR$ RealSense SR$300$ $3$D camera to obtain the facial depth map and adopt the same CNN classifier as in our multi-task network for SL$3$D.
In contrast, our method only utilizes the ordinary RGB camera \textit{without} specific hardware design.
To comprehensively compare our method with SL$3$D, we only select and perform the hardest presentation attacks that could cause failure cases of $2$D texture based methods.
The results in Tab.~\ref{3D-anti-spoofing} indicate that our method can achieve comparable anti-spoofing performance compared to SL$3$D.

\begin{table}[t!]
\centering
\setlength{\abovecaptionskip}{2mm}
\caption{\small $FAR$ comparisons with SL$3$D.
For each type of spoofing, we attack the system $100$ times and count the passing cases.
}
\label{3D-anti-spoofing}
\vspace{-2mm}
    \begin{tabular}{| c | c | c | c | c |}
    \hline
    Spoofing Type                       & SL3D          & Ours-AG       \\ \hline
    Paper Photo~(rect)               & 0/100          & 0/100             \\ \hline
    Paper Photo~(crop,twist)  & 1/100          & 0/100        \\ \hline
    Paper Photo~(crop,carve)     & 1/100          & 1/100     \\ \hline
    Screen Video~(iPad)            & 0/100          & 0/100         \\ \hline
    Projector Spoof                           & 0/100          & 0/100            \\ \hline
    {FAR(\%)}               & \textbf{0.40}          & \textbf{0.20}      \\ \hline
    \end{tabular}
\end{table}

\section{Conclusion}
In this paper, an effective facial anti-spoofing method named Aurora Guard is proposed, which holds real-time cross-platform applicability.
The key novelty of our method is to leverage two kinds of auxiliary information, the depth map and the light CAPTCHA based on light reflection, which significantly improve the accuracy and reliability of anti-spoofing system against unlimited $2$D presentation attacks.
Extensive experiments on public benchmark and our dataset show that AG is superior to the state of the art methods.

{\small
\bibliographystyle{named}
\bibliography{ijcai19}
}

\end{document}